\documentclass[twocolumn,twoside]{IEEEtran}

\usepackage{amsmath,amstext,amssymb,epsf}

\newcommand{\pair}[1]{\langle #1\rangle}

\newtheorem{theorem}{\sc Theorem}

\newtheorem{lemma}{\sc Lemma}
\newtheorem{coro}{\sc Corollary}
\newtheorem{req}{\sc Requirement}
\newtheorem{nota}{\sc Notation}
\newtheorem{defin}{\sc Definition}
\newtheorem{rem}{\sc Remark}
\newtheorem{cla}{\sc Claim}
\newtheorem{ex}{\sc Example}

\newenvironment{remark}{\begin{rem}}{\hspace*{\fill}$\Diamond$\end{rem}}

\newenvironment{definition}{\begin{defin}}{\end{defin}}

\renewcommand{\emptyset}{\varnothing}

\title{Approximation of the Two-Part MDL Code}
\author{Pieter Adriaans\thanks{Adriaans is supported in part by Perot
Enterprises Netherlands and Syllogic.
Address:
Kruislaan 419, Matrix I,
1098 VA Amsterdam,
The Netherlands. Email:
{\tt pieter@robosail.com}.}
and 
Paul Vit\'anyi\thanks{Vit\'anyi is supported in part by the BSIK Project BRICKS
of the Dutch government and NWO, and by the
EU NoE PASCAL (Pattern Analysis, Statistical Modeling, and Computational Learning).
Address: 
CWI, Kruislaan 413,
1098 SJ Amsterdam, The Netherlands. 
Email: {\tt Paul.Vitanyi@cwi.nl}.}
}

\date{}
\begin{document}
\maketitle

\begin{abstract}
Approximation of the optimal two-part MDL code for given data,
through successive 
monotonically length-decreasing two-part MDL codes, has the following
properties: (i) computation of each step may take arbitrarily long;
(ii) we may not know when we reach the optimum, or whether we will
reach the optimum at all; (iii) the sequence
of models generated may not monotonically improve the goodness of fit; 
but (iv) the
model associated with the optimum has (almost) the best goodness of fit.      
To express the practically interesting goodness of fit of individual models
for individual data sets
we have to rely on
Kolmogorov complexity. 

{\em Index Terms}---
minimum description length, 
model selection,
MDL code,
approximation,
model fitness,
Kolmogorov complexity,
structure functions,
examples

\end{abstract}

\section{Introduction}
In machine learning pure applications of
MDL are rare, partially because of the difficulties one encounters
trying to define an adequate model code and data-to-model code,
and partially because of the operational difficulties that are poorly
understood. We
analyze aspects of both the power and the perils of 
MDL precisely and formally.  Let us first resurrect a familiar
problem from our childhood to illustrate some of the issues involved.

The process of solving a jigsaw puzzle involves an
\emph{incremental reduction of entropy}, and this
serves to illustrate the analogous features of
the learning problems which are the main issues of this work.
Initially, when the pieces come out
of the box they have a completely random ordering. Gradually we
combine pieces, thus reducing the entropy and increasing the order until the
puzzle is solved. In this last stage we have found a maximal
ordering. Suppose that Alice and Bob both start to solve two
versions of the same puzzle, but that they follow different
strategies. Initially,  Alice sorts all pieces according to
color, and Bob starts by sorting the pieces according to
shape. (For the sake of argument we assume that the
puzzle has no recognizable edge pieces.) The crucial insight,
shared by experienced puzzle aficionados, is that Alice's
strategy is efficient whereas Bob's strategy is not and is in fact even
worse than a random strategy. Alice's strategy is efficient,
since the probability that pieces with about the same color match is
much greater than the unconditional probability of a match.
On the other hand the information about the shape of the pieces
can only be used in a relatively late stage of the puzzle process.
Bob's effort in the beginning is a waste of time, because he must
reorder the pieces before he can proceed to solve the puzzle. This
example shows that if the solution of a problem depends on finding
a \emph{maximal} reduction of entropy this does not mean that
\emph{every} reduction of entropy brings us closer to the solution.
Consequently reduction of entropy is not in all cases a good strategy.

\subsection{Entropy Versus Kolmogorov Complexity}
Above we use ``entropy''
in the often used, but inaccurate, sense of ``measure of unorderedness
of an individual arrangement.'' However, entropy is a measure of
uncertainty associated with a random variable, here a set of arrangements
each of which has a certain probability of occurring.
The entropy of every individual arrangement is by definition zero.
To circumvent this problem, often the notion of ``empirical entropy''
is used, where certain features like letter frequencies of the individual
object are analyzed, and the entropy is taken with respect to the 
set of all objects having the same features. The result obviously
depends on the choice of what features to use: no features gives
maximal entropy and all features (determining the individual object
uniquely) gives entropy zero again. Unless one has knowledge of the
characteristics of a definite random variable producing 
the object as a typical outcome, this procedure gives arbitrary and
presumably meaningless, results. This conundrum arises since classical
information theory deals with random variables and the communication
of information. It does not deal with the information (and the complexity
thereof) in an individual object independent of an existing
(or nonexisting) random variable
producing it.
 To capture the latter notion precisely one has to use
``Kolmogorov complexity'' instead of ``entropy,'' and we will do so in our
treatment. For now, the ``Kolmogorov complexity'' of a file
is the number of bits in the ultimately compressed version of the file
from which the original can still be losslessly extracted by a fixed general
purpose decompression program.

\subsection{Learning by MDL}
Transferring the jigsaw puzzling insights to the general case
of learning algorithms using the minimum description length
principle (MDL), \cite{Ri83,BRY,Ri07},  we observe that although it
may be true that 
the maximal compression yields the best solution, 
it may still not be true that every incremental compression brings
us closer to the solution. Moreover, in the case of many MDL problems there
is a complicating issue in the fact that the maximal compression
cannot be computed.

More formally,
in constrained model selection the model is taken from a given
model class.  Using two-part MDL codes for the given data,
we assume that the shortest two-part code for the data,
consisting of the model code and the data-to-model code, yields the best
model for the data. To obtain the shortest code, a natural way is to
approximate it by a process of finding ever shorter candidate two-part
codes. 
Since we start with a finite two-part code, and
 with every new candidate two-part code we decrease the code length,
eventually we must achieve the shortest two-part code (assuming that 
we search through all two-part codes for the data). Unfortunately,
there are two problems: (i) the computation to find the next shorter
two-part code may be very long, and we may not know how long; and
(ii) we may not know when we have reached the shortest two-part code:
with each candidate two-part code there is the possibility that further
computation may yield yet a shorter one. But because of item (i)
we cannot a priori bound the length of that computation.
There is also the possibility that the algorithm
will never yield the shortest two-part code because it 
considers only part of the search space or gets trapped in
a nonoptimal two-part code.

\subsection{Results}

We show that for some MDL algorithms the sequence 
of ever shorter two-part codes for the data converges in a finite number
of steps to the best model. However, for every MDL algorithm
the intermediate
models may not convergence
monotonically in goodness. In fact, in the sequence of candidate two-part codes
converging to a (globally or locally)
 shortest, it is possible that the models involved
oscillate from being good to bad. 
Convergence is only monotone if the model-code parts
in the successive two-part codes are always the shortest (most compressed)
codes for the models involved. But this property cannot be
guaranteed by any effective method.
 
It is very difficult, if not impossible, to formalize the
goodness of fit of an individual model for individual data in
the classic statistics setting, which is probabilistic.
Therefore, it is impossible to express the practically important
issue above in those terms. 
Fortunately, new developments
in the theory of Kolmogorov complexity \cite{Ko74,VV02} 
make it possible to rigorously
analyze the questions involved, possibly involving noncomputable
quantities. But it is better to have a definite statement in a theory
than having no definite statement at all. 
Moreover, for certain algorithms (like Algorithm Optimal MDL in
Theorem~\ref{alg.mdl}) we can guarantee that they satisfy
the conditions required, even though these are possibly noncomputable. 
In Section~\ref{sect.dm} 
we review the necessary notions from \cite{VV02}, both 
in order that the paper is self-contained and the definitions and notations are
extended from the previously used singleton data to multiple data samples. 
Theorem~\ref{theo.recoding} shows
that the use of MDL will be approximately invariant under recoding of the data.
The next two sections contain the main results:
Definition~\ref{def.MDLalg} defines the notion of an MDL algorithm.
Theorem~\ref{alg.mdl} shows that there exists such an MDL algorithm that
in the (finite) limit results in an optimal model.
The next statements are about MDL algorithms in general, also the ones
that do not necessarily result in an optimal MDL code.
Theorem~\ref{theo.approxim} states a sufficient condition
for improvement of the randomness deficiency (goodness of fit)
of two consecutive length-decreasing MDL codes. 
This extends Lemma V.2 of the
\cite{VV02} (which assumes all programs are shortest) and
corrects the proof concerned.
The theory is applied and illustrated in Section~\ref{sect.single}:
Theorem~\ref{theo.fluctuate} shows by example 
that a minor violation of the sufficiency
condition in Theorem~\ref{theo.approxim} can result in worsening
the randomness deficiency (goodness of fit)
of two consecutive length-decreasing MDL codes.  
The special case of learning DFAs from positive examples is
treated in Section~\ref{sect.multi}. The main result shows, 
for a concrete and computable MDL code, that a decrease in the length 
of the two-part MDL code
does not imply a better model fit (see Section~\ref{sect.lmdl})
unless there is a sufficiently
large decrease as that required in
Theorem~\ref{theo.approxim} (see Remark~\ref{rem.smc}).

\section{Data and Model}\label{sect.dm}

Let $x,y,z \in {\cal N}$, where
${\cal N}$ denotes the natural
numbers and we identify
${\cal N}$ and $\{0,1\}^*$ according to the
correspondence
\[(0, \epsilon ), (1,0), (2,1), (3,00), (4,01), \ldots \]
Here $\epsilon$ denotes the {\em empty word}.
The {\em length} $|x|$ of $x$ is the number of bits
in the binary string $x$, not to be confused with the {\em cardinality}
$|S|$ of a finite set $S$. For example,
$|010|=3$ and $|\epsilon|=0$, while $|\{0,1\}^n|=2^n$ and
$|\emptyset|=0$.
Below we will use the natural numbers and the binary strings
interchangeably. Definitions,  notations, and facts we use about prefix codes,
self-delimiting codes, and Kolmogorov complexity, can be found
in \cite{LiVi97} and are briefly reviewed in Appendix~\ref{sect.prel}.
                                                                                
The emphasis is on binary sequences only for convenience;
observations in any alphabet can be encoded in binary in a way
that is theory neutral.
Therefore, we consider only data 
$x$ in $\{0,1\}^*$. 
In a typical statistical inference situation we are given 
a subset of 
$\{0,1\}^*$,
the data sample, and are required to infer
a model  for the data sample. 
Instead of $\{0,1\}^*$
we will consider 
$\{0,1\}^{n}$ for some fixed but arbitrarily large $n$.
\begin{definition}
\rm
A {\em data sample} $D$ is a subset of $\{0,1\}^n$. 
For technical convenience we want a model $M$ for $D$ to contain 
information about the cardinality of $D$. 
A {\em  model} $M$ has the form $M = M' \bigcup \{\#i\}$,
where $M' \subseteq \{0,1\}^n$
and $i \in \{0,1\}^n$.
We can think of $i$ as the $i$th binary string in $\{0,1\}^{n}$.
Denote the cardinalities by lower case letters:
\[
d = |D|, \; m = |M'|.
\]
If $D$ is a data sample and {\em $M$ is a model for $D$} then
$D \subseteq M' \subseteq M$, $M=M' \bigcup \{\#d\}$,
and we write $M \sqsupset D$ or $D \sqsubset M$. 
\end{definition}

Denote the {\em complexity
of a finite set} $S$ by
$K(S)$---the length (number of bits) of the
shortest binary program $p$ from which the reference universal
prefix machine $U$
computes a lexicographic listing of the elements of $A$ and then
halts.
That is, if $S=\{x_1 , \ldots , x_{d} \}$, the elements given
 in  lexicographic order, then
$U(p)= \langle x_1,\langle x_2, \ldots, \langle x_{d-1},x_d\rangle \ldots\rangle \rangle $.
The shortest program $p$,
or, if there is more than one such shortest program, then
the first one that halts in a standard dovetailed running of all programs,
is denoted by $S^*$. 

The {\em conditional complexity} $K(D \mid M)$ of $D \sqsubset M$
is the length (number of bits) of the
shortest binary program $p$ from which the reference universal
prefix machine $U$
from input $M$ (given as a list of elements)
  outputs $D$ as a lexicographically ordered
 list of elements
and halts.
We have
        \begin{equation}\label{eq57}
K(D  \mid  M)\le\log {m \choose d}+ O(1).
        \end{equation}
The upper bound follows by considering a self-delimiting code of $D$ 
given $M$
(including the number $d$ of elements in $D$), consisting of
a $\lceil\log {m \choose d}\rceil$ bit long index
of $D$ in the lexicographic ordering of the number of ways to choose
$d$ elements from $M'=M-\{\#d\}$.
This code is called the
\emph{data-to-model code}.
Its length quantifies the maximal ``typicality,'' or ``randomness,''
any data sample $D$ of $d$ elements can have with respect 
to model $M$ with $M \sqsupset D$.

\begin{definition}
\rm
The lack of typicality
of $D$ with respect to $M$
is measured by the amount by which $K(D \mid M)$
falls short of the length of the data-to-model code.
The {\em randomness deficiency} of $D \sqsubset M$ 
is defined by
      \begin{equation}\label{eq:randomness-deficiency}
\delta (D  \mid  M) = \log {m \choose d} - K(D  \mid  M),
      \end{equation}
for $D \sqsubset M$, and $\infty$ otherwise.
\end{definition}

The randomness deficiency can be a little smaller than 0, but not more than
a constant. 
If the randomness deficiency is not much greater than 0,
then there are no simple special properties that
single $D$ out from the majority of data samples of cardinality $d$
to be drawn from $M'=M-\{\#d\}$.
This is not just terminology: If $\delta (D  \mid  M)$ is small enough, 
then $D$ satisfies {\em all} properties of low Kolmogorov complexity
that hold for the majority of subsets of cardinality $d$ of $M'$. To be precise:
A {\em property} $P$ represented by $M$ is a
subset of $M'$, and we say that
$D$ satisfies property $P$ if $D$ is a subset of $P$. 
                                                                                
\begin{lemma}
Let $d,m,n$ be natural numbers, and let 
$D \subseteq M' \subseteq \{0,1\}^{n}$, 
$M=M' \bigcup \{\#d\}$,
$|D|=d, |M'|=m$, and let $\delta$ be a simple
function of the natural numbers to the real numbers,
that is, $K(\delta)$ is a constant, 
for example,  $\delta$ is $\log$ or $\sqrt{}$.
                                                                                
(i) If $P$ is a property satisfied by all $D \sqsubset M$ with
$\delta(D  \mid  M) \le \delta (n)$,
then $P$ holds for a fraction of at
least $1-1/2^{\delta(n)}$ of the subsets of $M' = M-\{\#d\}$.
                                                                                
(ii) Let 
$P$ be a
property
that holds for a fraction of at least
$1-1/2^{\delta (n)}$ of the
subsets of $M'=M-\{\#d\}$.
There is a constant $c$, such that $P$ holds
for every $D \sqsubset M$
with $\delta (D  \mid  M)\le\delta (n)-K(P \mid M) -c$.
\end{lemma}
                                                                                
\begin{proof}
(i) By assumption, all data samples $D \sqsubset M$ 
with  
\begin{equation}\label{eq.fraction}
K(D|M) \geq \log {m \choose d} - \delta (n)
\end{equation}
 satisfy $P$.
There are only  
\[
\sum_{i=0}^{\log {m \choose d} - \delta (n)-1}2^i
= {m \choose d} 2^{- \delta (n)}-1
\]
programs of length smaller than $\log {m \choose d} - \delta (n)$,
so there are at most that many $D \sqsubset M$ 
that do not satisfy \eqref{eq.fraction}.
There are ${m \choose d}$ sets $D$ that satisfy $D \sqsubset M$,
and hence a fraction of at least $1-1/2^{\delta(n)}$ of
them satisfy \eqref{eq.fraction}.
                                                                                
(ii)
Suppose $P$ does not hold for a data sample $D \sqsubset M$
and the randomness deficiency \eqref{eq:randomness-deficiency} satisfies
$\delta(D| M) \leq \delta (n) -K(P|M)-c$.
Then we can reconstruct $D$ from a description of $M$,
and $D$'s index $j$ in an effective enumeration of all subsets
of $M$ of cardinality $d$ for
which $P$ doesn't hold. There are at
most ${m \choose d} /2^{ \delta (n)}$ such
data samples by assumption, and therefore there are constants 
$c_1,c_2$ such that
\[ K(D \mid M) \leq  \log j+ c_1 \leq \log {m \choose d} - \delta ( n) + c_2. \]
Hence, by the assumption on the randomness deficiency of
$D$, we find $K(P|M) \leq c_2 -c$,
which contradicts the necessary nonnegativity
of $K(P|M)$ if we choose $c > c_2$.
\end{proof}

The {\em minimal randomness deficiency} function 
of the data sample $D$ is defined by
           \begin{equation}
\label{eq1} 
\beta_D( \alpha) =
\min_{M} \{ \delta(D \mid  M): M \sqsupset D , \;  K(M) \leq \alpha \},
            \end{equation}
where we set $\min \emptyset = \infty$.
The smaller $\delta(D  \mid  M)$ is, the more $D$ can be considered
as a {\em typical} data sample
from $M$. This means that a set $M$ for which $D$ incurs minimal
randomness deficiency, in the model class of contemplated sets of given maximal
Kolmogorov complexity, is a ``best fitting'' model
for $D$ in that model class---a most likely explanation, and $\beta_D(\alpha)$
can be viewed as a {\em constrained best fit estimator}.

\subsection{Minimum Description Length Estimator}
The length of the minimal two-part code for $D$
with model $M \sqsupset D$ consist
of the model cost $K(M)$ plus the
length of the index of $D$ in the enumeration of choices of $d$ elements
out of $m$ ($m=|M'|$ and $M'=M-\{\#d\}$). 
Consider the model class of $M$'s of given maximal Kolmogorov 
complexity $\alpha$.
The {\em MDL} function or {\em constrained MDL estimator} is
  \begin{equation}\label{eq.3}
   \lambda_{D}(\alpha) =
\min_{M} \{\Lambda(M): M \sqsupset D,\; K(M) \leq \alpha\},
  \end{equation}
where $\Lambda(M)=K(M)+\log {m \choose d} \ge K(D)+O(1)$ is
the total length of two-part code of $D$
with help of the model $M$.
This function $\lambda_D (\alpha)$ is the
celebrated optimal two-part MDL code
length as a function of $\alpha$,
with the model class restricted to models
of code length at most $\alpha$. The functions $\beta_D$ and $\lambda_D$
are examples of Kolmogorov's {\em structure functions}, \cite{Ko74,VV02}.

Indeed,
consider the following \emph{two-part code}
for $D \sqsubset M$: the first part is
a shortest self-delimiting program $p$ for $M$ and the second
part is
$\lceil\log {m \choose d}\rceil$ bit long index of $D$
in the lexicographic ordering of all choices of $d$ elements from $M$.
Since $M$ determines $\log {m \choose d}$ this code 
is self-delimiting
and we obtain the two-part code,
where the constant $O(1)$ is
the length of an additional program that reconstructs
$D$ from its two-part code.
Trivially, $\lambda_D (\alpha) \geq K(D)+O(1)$.
For those $\alpha$'s that 
have
$\lambda_D (\alpha) = K(D)+O(1)$, the associated model $M \sqsupset D$ 
in at most $\alpha$ bits
(witness for
$\lambda_D(\alpha)$)
is called a {\em sufficient statistic} for $D$. 
\begin{lemma}
 If $M$ is a sufficient
statistic for $D$, then
the randomness deficiency of $D$ in $M$ is $O(1)$,
that is, $D$ is a typical
data sample for $M$, and $M$ is a model of best fit for $D$.
\end{lemma}
\begin{proof}
If $M$ is a sufficient
statistic for $D$, then $K(M)+\log {m \choose d} = K(D)+O(1)$. The left-hand
side of the latter equation
is a two-part description of $D$ using the model $M \sqsupset D$ and
as data-to-model code the index of $D$ in the enumeration
of the number of choices of $d$ elements from  $M$ in 
$\log  {m \choose d}$ bits. 
This left-hand side equals the right-hand side which
 is the shortest one-part
code of $D$ in $K(D)$ bits. Therefore, 
\begin{align*}
K(D) &\leq K(D,M) +O(1)
\\&\leq K(M)+K(D  \mid M)+O(1) 
\\& \leq K(M)+\log {m \choose d}+O(1) = K(D)+O(1).
\end{align*}
The first and second inequalities are straightforward, the third inequality 
states that given $M \sqsupset D$ 
we can describe $D$ in a self-delimiting manner in 
$\log  {m \choose d}+O(1)$ bits,
and the final equality follows by the sufficiency property. 
This sequence of (in)equalities implies
that $K(D \mid M)=\log  {m \choose d} +O(1)$. 
\end{proof}
\begin{remark}[Sufficient but not Typical]
\rm
Note that the  data sample $D$ can have randomness deficiency about 0, and
hence be a typical element
for models $M$, while $M$ is not a sufficient statistic.
A sufficient statistic $M$
for $D$ has the additional property, apart from being a model
of best fit, that $K(D,M)=K(D)+O(1)$
and therefore by \eqref{eq.soi} in Appendix~\ref{sect.prel}
we have $K(M|D^*)=O(1)$:
the sufficient statistic $M$ is a model of best fit
that is almost completely determined by $D^*$, a shortest program
for $D$.
\end{remark}
\begin{remark}[Minimal Sufficient Statistic]
\rm
The sufficient
statistic associated with $\lambda_D(\alpha)$
with the least $\alpha$ is called the
{\em minimal sufficient statistic}.
\end{remark}

\begin{remark}[Probability Models]
\rm
Reference \cite{VV02} and this paper analyze a canonical setting
where the models are finite sets. 
We can generalize the treatment to the case 
where the models are the computable
probability mass functions. The computability
requirement does not seem very restrictive.
We cover most, if not all,
probability mass functions ever considered, 
provided they have computable parameters.
In the case of multiple data we consider probability mass functions $P$ 
that map subsets $B \subseteq \{0,1\}^n$ into $[0,1]$ such that 
$\sum_{B \subseteq \{0,1\}^n} P(B) = 1$. For every $0 \leq d \leq 2^n$,
we define $P_d (B) = P(B \mid |B|=d)$.
For data $D$ with $|D|=d$ we 
obtain 
 $\lambda_D (\alpha) = \min_{P_d} \{K(P_d)+ \log 1/P_d(D): 
P_d(D) > 0$ and $P_d$ is a 
computable probability mass function with $K(P_d) \leq \alpha$$\}$. 
The general
model class of computable probability mass functions is equivalent to
the finite set model class, up to an additive logarithmic $O( \log dn)$
term. This result for multiple data
 generalizes the corresponding result for singleton data in \cite{Sh83,VV02}. 
Since the other results in \cite{VV02} such as \eqref{eq.eq}
and those in Appendix~\ref{sect.formal}, generalized to multiple data,
 hold only up to
 the same additive logarithmic
term anyway, they carry over to the probability models.
\end{remark}

The generality of the results are at the same time a restriction.
In classical statistics one is commonly interested in model classes
that are partially poorer and partially richer than the ones we consider.
For example, the class of Bernoulli processes, or $k$-state Markov
chains, is poorer than the class of computable probability mass functions
of moderate maximal Kolmogorov complexity $\alpha$,
in that the latter class may contain
functions that require far more complex computations than the rigid
syntax of the classical classes allows. Indeed, the class of computable
probability mass functions of even moderate complexity allows
implementation of a function mimicking a universal Turing machine computation.
On the other hand, even the simple Bernoulli process can be equipped
with a noncomputable real bias in $(0,1)$, and hence the generated
probability mass function over $n$ trials is not a computable function.
This incomparability of the algorithmic model classes studied here and
the traditional statistical model classes, means that the
current results cannot be directly transplanted to the traditional setting.
They should be regarded as pristine truths that hold in a
platonic world that can be used as guideline to develop analogues
in model classes that are of more traditional concern, as in
\cite{Ri07}.

\subsection{Essence of Model Selection }
\label{sect.essence}
The first parameter we are interested in is the {\em simplicity}
$K(M)$ of the
model $M$ explaining the data sample $D$ ($D \sqsubset M$).
The second parameter is  
{\em how typical} the data is
with respect to $M$, expressed by
the randomness deficiency
$\delta(D \mid M)=\log {m \choose d}-K(D  \mid M)$.
The third parameter is  
how {\em short
the two part code}
$\Lambda(M)=K(M )+\log {m \choose d}$
of the data sample $D$ using theory $M$ with $D \sqsubset M$ is. 
The second part consists of the full-length index,
ignoring saving in code length using possible nontypicality
of $D$ in $M$ (such as being the first $d$ elements in the enumeration of 
$M'=M-\{\#d\}$).
These parameters induce a partial order on the contemplated set of models.
We write 
$M_1 \le M_2$,  if $M_1$ scores equal or less than
$M_2$ in all three
parameters. If this is the case, then we may say that
$M_1$ is at least as good as $M_2$
as an explanation for $D$ (although the converse need not necessarily hold,
in the sense that it is possible that $M_1$ is
at least as good a model for $D$ 
as $M_2$ without
scoring better than $M_2$ in all three parameters simultaneously).

The algorithmic statistical properties of a data sample $D$ are
fully represented by
the set $A_D$ of all
triples
\[
\pair{ K(M), \delta(D \mid M), \Lambda(M) }
\]
with $M \sqsupset D$, together with a component wise
order relation on the elements of those triples.
The complete characterization of
this set 
follows from
the results in \cite{VV02}, provided we generalize the singleton case treated
there to the multiple data case required here.

In that reference it is shown
that 
if we minimize the length of a two-part code for an individual data sample,
the two-part code consisting of
a model description and a data-to-model code
over the {\em class of all computable models} of at most a given complexity,
then the following is the case.
With {\em certainty}
and not only with high probability as in the classical case 
this process selects an individual model that
in a rigorous sense is (almost) 
the best explanation for the individual data sample
that occurs  among the contemplated models.
(In modern versions of MDL, \cite{Gr07,BRY,Ri07}, one
    selects the model that
    minimizes just the data-to-model code length
    (ignoring the model code length), or minimax and mixture MDLs. 
These are not treated here.)
These results are exposed in the proof and analysis of the 
equality:
         \begin{equation}\label{eq.eq}
\beta_D (\alpha )   = \lambda_D (\alpha)
- K(D),
         \end{equation}
which holds within negligible additive $O (\log dn)$ terms, 
in argument and value. We give the precise statement in
\eqref{eq.multipleeq} in  Appendix~\ref{sect.formal}.
\begin{remark}\label{rem.witness}
\rm
Every model (set) $M$ that witnesses the value 
$\lambda_D(\alpha)$,
also witnesses the value $\beta_D (\alpha)$ (but not vice versa).
The functions $\lambda_D$ and $\beta_D$ can assume all
possible shapes over their full domain of definition (up to
additive logarithmic precision in both argument and value).
We summarize these matters in Appendix~\ref{sect.formal}.
\end{remark}

\subsection{Computability}
\label{sect.comp}
How difficult is it to compute the functions $ \lambda_D, \beta_D$,
and the minimal sufficient statistic? To express the properties
appropriately we require the notion of functions
that are not computable,
but can be approximated monotonically by a computable
function.
\begin{definition}
\rm
\label{def.semi}
A function $f: {\cal N} \rightarrow {\cal R}$ is
{\em upper semicomputable} if there is a Turing machine $T$ computing a
total function $\phi$
such that $\phi (x,t+1) \leq \phi (x,t)$ and
$\lim_{t \rightarrow \infty} \phi (x,t)=f(x)$. This means
that $f$ can be computably approximated from above.
If $-f$ is upper semicomputable, then $f$ is lower semicomputable.
A function is called {\em semicomputable}
if it is either upper semicomputable or lower semicomputable.
If $f$ is both upper semicomputable and lower semicomputable,
then we call $f$ {\em computable} (or recursive if the domain
is integer or rational).
\end{definition}

To put matters in perspective: even if a function is computable,
the most feasible type identified above, this doesn't mean much in
practice. Functions like $f(x)$ of which the computation terminates
in computation time of 
$t(x) = x^x$ (say measured in flops), are among the easily computable ones. 
But for $x=30$, even a computer performing
an unrealistic Teraflop per second, 
requires $30^{30}/ 10^{12} > 10^{28}$ seconds.
This is more than $3 \cdot 10^{20}$ years. It is out of the question
to perform such computations. Thus, the fact that a function
or problem solution is computable gives no insight in how {\em feasible}
it is. But there are worse functions and problems possible: For example,
the ones that are semicomputable but not computable. Or worse yet,
functions that are not even semicomputable.

Semicomputability gives no knowledge of convergence guarantees: even though
the limit value is monotonically approximated, at no stage
in the process do we know how close we are to the limit value.
In Section~\ref{ex.MDL}, the indirect method  of Algorithm Optimal MDL shows 
that the function $\lambda_D$ (the MDL-estimator)
can be monotonically approximated
in the upper semicomputable sense. 
But in \cite{VV02} it was shown for singleton data samples,
and therefore {\em a fortiori} for multiple data samples $D$,
the fitness function $\beta_D$ (the direct method of Remark~\ref{rem.direct})
cannot be monotonically approximated in that sense, nor in the
lower semicomputable sense, in both cases not even
up to any relevant precision. Let us formulate this a little more
precisely:

The functions $ \lambda_D (\alpha), \beta_D(\alpha)$ 
have a finite domain
for a given $D$ and hence can be given as a table---so formally speaking
they are computable. But this evades the issue: there is no
algorithm that computes these functions for given $D$ and $\alpha$.
Considering them as two-argument functions it was
shown (and the claimed precision quantified):
\begin{itemize}
\item The function $\lambda_D(\alpha)$
is upper semicomputable but not computable up to any reasonable
precision.
\item  There is no algorithm that given $D^*$ and $\alpha$ finds
$\lambda_D(\alpha)$.
\item  The function $\beta_D(\alpha)$
is not upper nor lower semicomputable, not even to any reasonable precision.
To put $\beta_D(\alpha)$'s computability properties in perspective, 
clearly we can compute it given an oracle for the halting
problem. 
\begin{quote}
The {\em halting problem} is the problem 
whether an arbitrary Turing machine 
started on an initially all-0 tape will eventually terminate or
compute forever. This problem was shown to be undecidable by A.M. Turing
in 1937, see for example
\cite{LiVi97}. An oracle for the halting problem will, when asked, tell
whether a given Turing machine computation will or will not terminate.
Such a device is assumed in order to
determine theoretical degrees of (non)computability, and
is deemed not to exist.
\end{quote}
But using such an oracle gives us power beyond effective (semi)computability
and therefore brings us outside the concerns of this paper.
\item There is no algorithm
that given $D$ and $K(D)$ finds a minimal sufficient statistic for $D$
up to any reasonable precision.
\end{itemize}

\subsection{Invariance under Recoding of Data}
\label{ex.recoding}
In what sense are the functions invariant
under recoding of the data? If the functions $\beta_D$ and  $\lambda_D$
give us the stochastic properties of the data $D$, then we would not expect
those properties to change under recoding of the data into another format.
For convenience, let us look at a singleton example. 
Suppose we recode $D= \{x\}$ 
by a shortest program $x^*$ for it.
Since $x^*$ is incompressible
it is a typical element of the set of all strings of length $|x^*|=K(x)$,
and hence $\lambda_{x^*} (\alpha)$ drops to the Kolmogorov complexity $K(x)$
already for some $\alpha \leq K(K(x))$, so almost immediately (and it
stays within logarithmic distance of that line henceforth).
That is,
$\lambda_{x^*} (\alpha) = K(x)$ up to
logarithmic additive terms in argument and value,
irrespective of the (possibly quite different)
shape of $\lambda_x$. Since the Kolmogorov complexity function
$K(x)=|x^*|$ is not recursive, \cite{Ko65},
the recoding function $f(x) = x^*$ is also not recursive.
Moreover,  while $f$ is one-to-one and total
it is not onto.
But it is the
partiality of the inverse function (not all strings are shortest
programs) that causes the collapse of the structure function.
If one restricts the finite sets containing $x^*$ to be subsets of
$\{y^*: y \in \{0,1\}^n\}$, then the resulting
function $\lambda_{x^*}$ is within a logarithmic strip around $\lambda_x$.
The coding function $f$ is upper semicomputable and deterministic. 
(One can consider
other codes, using more powerful computability assumptions or probabilistic
codes, but that is outside the scope of this paper.) 
However, the structure function
is invariant under ``proper'' recoding of the data.
                                                                                
\begin{theorem}\label{theo.recoding}
Let $f$ be a recursive permutation of the set of 
finite binary strings in $\{0,1\}^n$
(one-to-one, total, and onto), and extend $f$ to subsets $D \subseteq \{0,1\}^n$.
 Then,
$\lambda_{f(D)}$ is ``close'' to  $\lambda_D$ in the sense that the graph of
$\lambda_{f(D)}$ is situated within a strip of width $K(f)+O(1)$ around 
the graph of $\lambda_D$.
\end{theorem}
\begin{proof}
Let $M \sqsupset D$ be a witness of $\lambda_D(\alpha)$. Then,
$M_f = \{f(y): y \in M\}$ satisfies $K(M_f) \leq \alpha + K(f)+O(1)$
and $|M_f|=|M|$. Hence, $\lambda_{f(D)} (\alpha + K(f)+O(1)) \leq \lambda_D(\alpha)$.
Let $M^f \sqsupset f(D)$ be a witness of $\lambda_{f(D)} (\alpha)$. Then,
$M^f_{f^{-1}} = \{f^{-1} (y): y \in M^f\}$ satisfies
$K(M^f_{f^{-1}}) \leq \alpha + K(f)+O(1)$ and $|M^f_{f^{-1}}|=|M^f|$.
Hence, $\lambda_{D} (\alpha + K(f)+O(1)) \leq \lambda_{f(D)}(\alpha)$ (since
$K(f^{-1}) = K(f)+O(1)$).
\end{proof}

\section{Approximating the MDL Code}
\label{ex.MDL}
   Given  $D\subseteq \{0,1\}^n$, 
the data to explain, and the
   model class consisting of all models $M$ for $D$
that have complexity $K(M)$ at most 
  $\alpha$. This $\alpha$ is
   the maximum complexity of an explanation we allow. 
As usual, we denote $m=|M|-1$ (possibly indexed like $m_t = |M_t|-1$)
and $d=|D|$. We search for programs $p$
   of length at most $\alpha$ that print a finite set $M\sqsupset  D$. Such
   pairs $(p,M)$ are possible explanations.
   The {\em best explanation} is defined to be
   the $(p,M)$ for
   which $\delta(D \mid M)$ is minimal, that is, 
$\delta(D \mid M)=\beta_D(\alpha)$. Since the function
   $\beta_D(\alpha)$ is not computable, there is no algorithm that halts with
   the best explanation.
   To overcome this problem
   we minimize the randomness deficiency by minimizing the MDL code
   length, justified by \eqref{eq.eq}, and thus maximize the
   fitness of the model for this data sample. Since  \eqref{eq.eq} holds only
up to a small error we should more 
properly say ``almost  minimize the randomness deficiency''
and ``almost  maximize the
   fitness of the model.'' 
\begin{definition}\label{def.MDLalg}
\rm
An algorithm $A$ is an {\em MDL algorithm} if the following holds.
Let $D$ be a data sample consisting of $d$ separated words of length $n$
in $dn+O(\log dn)$ bits. 
Given inputs $D$ and $\alpha$ ($0 \leq \alpha \leq dn +O(\log dn)$), 
algorithm $A$
written as $A(D, \alpha )$ produces a finite sequence of
pairs $(p_1,M_1), (p_2,M_2), \ldots , (p_{f}, M_f)$, such that 
every $p_t$ is a binary program
of length at most $\alpha$ that
prints a finite set 
$M_t$
with $D \sqsubset M_t$ and
 $|p_t|+\log {{m_t} \choose d} < |p_{t-1}|+\log {{m_{t-1}} \choose d}$  for
every  $1 < t \leq f$.
\end{definition}
\begin{remark}
\rm
It follows that 
 $K(M_t) \leq |p_t|$ for all $1 < t \leq f$.
Note that an MDL algorithm may consider only
a proper subset of all binary programs of length at most $\alpha$. In particular,
the final $|p_f|+\log {{m_f} \choose d}$ may be greater than
the optimal MDL code of length
$\min \{ K(M)+\log {{m} \choose d}: M \sqsupset D, \; K(M) \leq \alpha \}$. 
This happens when a program
$p$ printing $M$  with $ M \sqsupset D$ and $|p|= K(M) \leq \alpha$ is not
in the subset of binary programs considered by the algorithm, or the
algorithm gets trapped in a suboptimal solution.
\end{remark}

The next theorem gives an MDL algorithm that always finds the optimal
MDL code and, moreover, the model concerned 
is shown to be an approximately best fitting
model for dat $D$.

\begin{theorem}\label{alg.mdl}
There exists an MDL algorithm which given $D$ and $\alpha$
satisfies
$\lim_{t \rightarrow \infty} (p_t,M_t) = (\hat{p},\hat{M})$,
such that $\delta(D|\hat{M}) \leq \beta_D(i-O(\log dn))+O(\log dn)$.
\end{theorem}
\begin{proof}
We exhibit such an MDL algorithm:

{\bf Algorithm Optimal MDL ($D,\alpha$)}
\begin{description}
\item{\bf Step 1.}
Let $D$ be the data sample.
   Run all binary
   programs $p_1,p_2, \ldots$ of length at most $ \alpha$ in 
lexicographic length-increasing
order in a dovetailed style.
 The computation proceeds by stages $1,2, \ldots ,$
   and in each stage $j$ the overall computation executes step $j-k$
   of the particular subcomputation of $p_k$,
   for every $k$ such that $j-k >0$.
\item{\bf Step 2.}
   At every computation step $t$,
  consider all pairs $(p,M)$ such that
   program $p$ has printed the set $M \sqsupset D$ by time $t$.
   We assume that there is a first elementary computation step 
$t_0$ such that there is such a pair.
   Let a {\em best explanation} $(p_t,M_t)$ at computation step $t \geq t_0$ be
   a pair that minimizes the sum
   $|p|+\log {m \choose d}$ among all
   the pairs $(p,M)$.
\item{\bf Step 3.}
   We only change the best explanation $(p_{t-1},M_{t-1})$  of
computation step $t-1$ to
$(p_{t},M_{t})$ at computation step $t$,
if $|p_t|+\log {{m_t} \choose d} < |p_{t-1}|+\log {{m_{t-1}} \choose d}$.
\end{description}
   In this MDL algorithm 
the best explanation $(p_t,M_t)$ changes from time to time
   due to the appearance of a strictly better explanation. 
Since no pair $(p,M)$ can be
elected as best explanation twice, and there are  only finitely
many pairs, from some moment onward the explanation
   $(p_t,M_t)$ which is declared best does not change anymore.
Therefore the limit $(\hat{p},\hat{M})$ exists.
The model $\hat{M}$ is a witness set of $\lambda_D(i)$. The lemma
follows by (\ref{eq.eq}) and Remark~\ref{rem.witness}. 
\end{proof}

Thus, if we continue to approximate the two-part MDL code contemplating
every relevant model, then we will eventually
reach the optimal two-part code whose associated model
 is approximately the best explanation. That
is the good news. The bad news is that we do not know
when we have reached
this optimal solution. The functions
$h_D$ and $\lambda_D$, and their witness sets, cannot be computed
within any reasonable accuracy, Section~\ref{sect.comp}. 
Hence, there does not
exist a criterion
we could use to terminate the approximation somewhere
close to the optimum.

  In the practice of the real-world MDL, in the
   process of finding the optimal two-part MDL code,
or indeed a suboptimal two-part MDL code,
   we often have to be satisfied
 with running times $t$ that are much less than the time to
   stabilization of the best explanation.
For such small $t$, the model
   $M_t$ has a weak guarantee of goodness, since we know that
\[
\delta(D|M_t) + K(D) \le |p_t|+\log {{m_t} \choose d},
\]
because $K(D) \leq K(D,M_t) \leq K(M_t)+K(D|M_t)$
and therefore $K(D)-K(D|M_t) \leq K(M_t)\leq |p_t|$ (ignoring additive
constants).
 That is,
   the randomness deficiency of $D$ in
$M_t$ plus $K(D)$ is less than the
   known value $|p_t|+\log {{m_t} \choose d}$.
   Theorem~\ref{alg.mdl} implies that
   Algorithm  MDL gives not only {\em some} guarantee of goodness
during the approximation process
(see Section~\ref{sect.comp}),
   but also that, in the limit, that guarantee  approaches the value
   of its lower bound, that is, $\delta(D|\hat{M}) + K(D)$.
 Thus, in the limit,
   Algorithm Optimal MDL will yield an explanation that is only a little
   worse than the best explanation.

\begin{remark}\label{rem.direct}
{\bf (Direct Method)}
\rm
Use the same dovetailing process as in Algorithm Optimal MDL, with the
following addition.
At every elementary computation step $t$,
   select a
   $(p,M)$ for which $\log {m \choose d}-K^t(D|M)$ is minimal
among all programs $p$ that
   up to this time have printed a set  $M \sqsupset D$.
   Here $K^t(D|M)$ is the approximation of  $K(D|M)$
   from above defined by
$K^t(D|M)=\min\{|q|:$ the reference universal prefix machine $U$ outputs
$D$  on input $(q,M)$
   in at most  $t$ steps$\}$. Hence, $\log {m \choose d}-K^t(D|M)$
is an approximation from below to $\delta (D|M)$.
   Let $(q_t,M_t)$ denote the best explanation after $t$ steps.
 We only change the best explanation at computation step $t$,
 if $\log {{m_t} \choose d} - K^t(D|M_t)
<\log {m_{t-1} \choose d} - K^{t-1}(D|M_{t-1})$.
  This time the same explanation
   can be
   chosen as the best one twice. However, from some time $t$ onward, the best explanation
   $(q_t,M_t)$ does not change anymore.
 In the approximation process, the model $M_t$
   has no guarantee of goodness at all:
Since $\beta_D(\alpha)$ is not semicomputable up 
to any significant precision, Section~\ref{sect.comp},
   we cannot know a significant 
upper bound neither for $\delta(D|M_t)$, nor for
   $\delta(D|M_t) + K(D)$.
   Hence, we must prefer the indirect method of Algorithm Optimal MDL, approximating
a witness set for $\lambda_D(\alpha)$, instead of the direct one of approximating
a witness set for $\beta_D(\alpha)$.
\end{remark}

\section{Does Shorter MDL Code Imply Better Model?}
In practice we often must terminate an MDL algorithm as 
in Definition~\ref{def.MDLalg} prematurely.
A natural assumption is that the longer we
approximate the optimal two-part MDL code
the better the resulting model explains the data. Thus,
it is tempting to simply assume that in the approximation
every next shorter two-part MDL code also yields a better model.
However, this is not true.
To give an example
that shows where things go wrong
it is easiest to first give the conditions under
which premature search termination
is all right.
Suppose we replace
the currently best explanation
   $(p_1,M_1)$ in an MDL algorithm with explanation
   $(p_{2},M_{2})$ only if $|p_{2}|+\log {{m_{2}} \choose d}$
 is not just less than  $|p_1| +\log {{m_1} \choose d}$,
but less by more than the excess of $|p_1|$ over
$K(M_1)$.
 Then, it turns out that every time we change the explanation we improve
   its goodness.

   \begin{theorem}\label{theo.approxim}
Let $D$ be a data sample with $|D|=d$  ($0 <d<2^n$). Let
   $(p_1, M_1)$ and $(p_{2},M_{2})$ be
   sequential {\rm (}not necessary consecutive{\rm )} 
candidate best explanations.
produced by an MDL algorithm $A(D, \alpha)$.
   If
\begin{eqnarray*}
|p_{2}|+ \log {{m_{2}} \choose d} &\leq & |p_1|  +  
\log {{m_1} \choose d} 
\\&& - (|p_1|-K(M_1)) 
   - 10 \log \log {{2^n} \choose d} ,
\end{eqnarray*}
   then
$
   \delta (D | M_{2}) \le \delta (D | M_1) - 5 \log  \log {{2^n} \choose d}.
$
   \end{theorem}
   \begin{proof}
   For every pair of sets $M_1,M_{2} \sqsupset D$ we have
   \[
\delta (D | M_{2} ) - \delta (D | M_1) =
    \Lambda +  \Delta,
\]
with $\Lambda = \Lambda (M_{2}) - \Lambda (M_1) $ and
\begin{eqnarray*}
\Delta & =& -K(M_{2})-K(D|M_{2})  + K(M_1) + K(D|M_1) 
\\& \leq & - K(M_2,D) + K(M_1,D) + K(M_1^*|M_1) +O(1) 
\\&\leq&K(M_1,D|M_{2},D)+ K(M_1^*|M_1)+O(1).
\end{eqnarray*}
The first inequality uses the trivial
$-K(M_2,D) \geq -K(M_2)-K(D|M_2)$ and the nontrivial
$ K(M_1,D) + K(M_1^*|M_1) \geq K(M_1) + K(D|M_1)$ which follows by
\eqref{eq.soi}, and the second inequality uses the general property that
 $K(a|b) \geq K(a)-K(b)$.
By the assumption in the theorem,
\begin{eqnarray*}
\Lambda & \le &
   |p_{2}|+\log {{m_{2}} \choose d}- \Lambda (M_1) 
\\ &=&|p_{2}|+\log {{m_{2}} \choose d}- \left( |p_1| 
  + \log{{m_1} \choose d} \right) 
\\ &&    + (|p_1|-K(M_1)) 
\\& \le & 
   - 10 \log \log { {2^n} \choose d}.
\end{eqnarray*}
Since by assumption the difference in MDL codes 
$\Lambda = \Lambda (M_2) - \Lambda (M_1) > 0$, 
it suffices to show that
   $K(M_{1},D | M_2,D)+K(M_1^*|M_1) \le 5 \log \log {{2^n} \choose d}$
to prove the theorem.
   Note that $(p_1,M_1)$ and $(p_{2},M_{2})$ are in this order 
sequential candidate
best explanations
   in the algorithm, and every candidate best explanation may appear only once.
   Hence, to identify $(p_1,M_1)$ we only need to know the MDL algorithm $A$,
the maximal complexity $\alpha$ of the contemplated models, the data sample $D$,
   the candidate explanation $(p_{2},M_{2})$,
and the number $j$ of candidate best explanations in between
 $(p_1,M_1)$ and $(p_{2},M_{2})$.
To identify $M_1^*$ from $M_1$ we only require $K(M_1)$ bits.
The program $p_{2}$ can be found from $M_{2}$ and
   the length $|p_{2}| \leq \alpha$, as the first program computing $M_{2}$
   of length $|p_{2}|$ in the process of running
the algorithm $A(D, \alpha)$.
Since $A$ is an MDL algorithm we have $j \leq |p_1| + 
\log {{m_1} \choose d} \leq \alpha+ \log {{2^n} \choose d}$,
 and $K(M_1) \leq \alpha$. Therefore,
   \begin{eqnarray*}
&&K(M_{1},D | M_2,D)+ K(M_1^*|M_1) 
\\ &&\le  \log |p_{2}|+\log \alpha
+ \log K(M_i) +  \log j + b
\\&& \le  3\log \alpha+ \log \left(\alpha+ \log {{2^n} \choose d} \right) + b, 
\end{eqnarray*}
where $b$ is the number of bits we need to encode
the description of the
MDL algorithm,
the descriptions of the constituent
parts self-delimitingly,
and the description of a program to reconstruct $M_1^*$ from $M_1$.
Since $\alpha \leq n+O(\log n)$, we find
\begin{eqnarray*}
&&K(M_{1},D | M_2,D) +K(M_1^*|M_1) 
\\&&\leq 3 \log n + \log \log {{2^n} \choose d} +
O\left(\log \log  \log {{2^n} \choose d}\right)
\\&&\leq 5 \log \log {{2^n} \choose d},
\end{eqnarray*}
where the last inequality follows from $0 < d < 2^n$ and $d$
being an integer.
   \end{proof}
\begin{remark}
\rm
We need an MDL algorithm in order to restrict the 
sequence of possible candidate models examined 
to at most $\alpha + \log {{2^n} \choose d}$ with 
$\alpha \leq nd +O(\log nd)$ rather than all of the $2^{2^n-d}$  
possible models $M$ satisfying $M \sqsupset D$.
\end{remark}
\begin{remark}
\rm
In the sequence $(p_1,M_1), (p_2,M_2), \ldots ,$
of candidate best explanations produced by an MDL algorithm,
$(p_{t'},M_{t'})$ is actually better than
$(p_{t},M_{t})$ ($t < t'$), if
the improvement in the two-part MDL code-length is
the given logarithmic term in excess of
the unknown, and in general noncomputable
$|p_t|-K(M_t)$.
On the one hand, if
$|p_t|=K(M_t)+O(1)$, and
\[
|p_{t'}|+ \log  {{m_{t'}} \choose d} 
\leq |p_t| + \log  {{m_t} \choose d} - 10 \log \log {{2^n} \choose d},
\]
then $M_{t'}$ is a better explanation for data sample $D$
than $M_t$, in the sense that
   \[
\delta (D | M_{t'}) \le \delta (D | M_t) - 5 \log \log {{2^n} \choose d}.
\]
On the other hand, if $|p_t| - K(M_t)$ is large, 
then $M_{t'}$ may be a
much worse explanation than $M_t$.
Then, it is possible that
we improve the two-part MDL code-length by giving a worse model
$M_{t'}$ using, however, a $p_{t'}$
such that $|p_{t'}|+ \log {{m_{t'}} \choose d} 
< |p_{t}|+ \log {{m_{t}} \choose d}$ while
$\delta (D | M_{t'}) > \delta (D | M_t)$.
\end{remark}

\section{Shorter MDL Code May Not Be Better} 
\label{sect.single}
Assume that we want to infer
a language, given a single positive example (element of the language).
The positive example is $D=\{x\}$ with
$x = x_1 x_2 \ldots x_n$, $x_i \in \{0,1\}$ for $1 \leq i \leq n$.
We restrict the question to inferring
a language consisting of a set of elements of the same
length as the positive example, that is,
we infer a subset of $\{0,1\}^n$. We can view this as inferring the slice $L^n$
of the (possibly infinite) 
target language $L$ consisting of all words of length $n$ in the target
language. We identify the singleton data 
sample $D$ with its constituent
data string $x$. For the models we always have $M=M' \bigcup \{\#1\}$ 
with $M' \subseteq \{0,1\}^n$.
For simplicity we delete the cardinality indicator $\{\#1\}$ since it is always
1 and write $M = M' \subseteq \{0,1\}^n$.

Every $M \subseteq \{0,1\}^n$ can be represented by its characteristic
sequence $\chi = \chi_1 \ldots \chi_{2^n}$ with $\chi_i =1$
if the $i$th element of $\{0,1\}^n$ is in $M$, and 0 otherwise.
Conversely, every string of $2^n$ bits is the characteristic sequence
of a subset of $\{0,1\}^n$. Most of these subsets are ``random'' in the sense
that they cannot be represented concisely: their characteristic sequence
is incompressible. Now choose some integer $\delta$.
 Simple counting tells us that there are
only $2^{2^n - \delta} -1$ binary strings of length $<2^n - \delta$.
Thus, the number of possible binary programs of length $<2^n - \delta$
is at most $2^{2^n - \delta} -1$. This in turn implies (since every
program describes at best one such set) that the number of 
subsets $M \subseteq \{0,1\}^n$ with $K(M|n) <2^n - \delta$ is at most
$2^{2^n - \delta} -1$. Therefore, the number of 
subsets $M \subseteq \{0,1\}^n$ with 
\[ 
K(M|n) \geq 2^n - \delta
\]
 is
greater than 
\[
(1- 1/2^{\delta})2^{2^n}. 
\]
Now if $K(M)$ is significantly
greater than  $K(x)$, then
it is impossible to learn $M$ from $x$. This follows already from
the fact that $K(M|x) \geq K(M|x^*)+O(1) 
= K(M)-K(x) + K(x|M^*)+O(1)$ by \eqref{eq.soi}
(note that $K(x|M^*) > 0$). That is, we need more than $K(M)-K(x)$
extra bits of dedicated information to deduce $M$ from $x$.
Almost all 
sets in $\{0,1\}^n$ have so high complexity that no 
effective procedure can infer this set from a single example.
This holds in particular for every (even moderately) random set.

Thus, to infer such a subset 
$M \subseteq \{0,1\}^n$, 
given a sample datum $x \in M$,
using the MDL principle is clearly out of the question.
The datum $x$ can be
literally described in $n$ bits by the trivial MDL code $M=\{x\}$
with $x$ literal at self-delimiting 
model cost at most $n+O(\log n)$ bits and data-to-model cost
$\log |M|=0$.
It can be concluded that the only sets
$M$ that can possibly be inferred from $x$ (using MDL or any other
effective deterministic procedure)
are those that have $K(M) \leq K(x) \leq n + O(\log n)$. Such sets
are extremely rare: only an at most 
\[
2^{-2^n+n+ O(\log n)}
\]
fraction of all subsets
of $\{0,1\}^n$ has that small prefix complexity. This negligible fraction of
possibly learnable sets shows that such sets are very nonrandom; 
they are simple in the
sense that their characteristic sequences
have great regularity  
(otherwise the Kolmogorov complexity could not be this small).  
But this is all right: we do not want to learn random, meaningless, languages,
but only languages that have meaning. ``Meaning'' is necessarily expressed in
terms of regularity. 

Even if we can learn the target
model by an MDL algorithm in the limit, by selecting a sequence of models
that decrease the MDL code with each next model, it can still
be the case that a later model
in this sequence is a worse model than a preceding one. 
Theorem~\ref{theo.approxim} showed conditions that prevent this from happening.
We now show that if those conditions are not satisfied, it can indeed happen.

\begin{theorem}\label{theo.fluctuate}
There is a datum $x$ ($|x|=n$) with
explanations
   $(p_t,M_t)$ and 
   $(p_{t'},M_{t'})$  such that
   $|p_{t'}|+\log m_{t'}\le|p_t|+\log m_t- 10 \log n$ 
but
$\delta (x|M_{t'}) \gg \delta (x|M_t)$.
That is, $M_{t'}$ is much worse fitting
than $M_t$.
There is an MDL algorithm $A(x, n)$ generating $(p_t,M_t)$ and
   $(p_{t'},M_{t'})$ as best explanations with $t' > t$.
\end{theorem}
\begin{remark}
\rm
Note that the condition of Theorem~\ref{theo.approxim} 
is different from the first inequality in Theorem~\ref{theo.fluctuate} since
the former required an extra $-|p_t|+K(M_t)$ term in
the right-hand side.
\end{remark}
\begin{proof}
Fix 
datum $x$ of length $n$ which can be divided in
$uvw$ with $u,v,w$ of equal length 
(say $n$ is a multiple of 3)
with
$K(x)=K(u)+K(v)+K(w)= \frac{2}{3}n$, 
$K(u)=\frac{1}{9}n$, $K(v)=\frac{4}{9}n$, and $K(w)=\frac{1}{9}n$ 
(with the last four
equalities holding up to additive $O(\log n)$
terms).
Additionally, take $n$ sufficiently large so that 
$0.1n \gg 10 \log n$.

Define $x^i=x_1 x_2 \ldots x_i$ and
an MDL algorithm $A(x,n)$ that
examines the sequence of models 
$M_i = \{x^i\} \{0,1\}^{n-i}$, with
$i=0, \frac{1}{3}n, \frac{2}{3}n, n$.
The algorithm starts with candidate model $M_0$ and switches
from the current candidate to candidate $M_i$, $i= \frac{1}{3}n, \frac{2}{3}n, n$,
if that model gives a shorter MDL code
than the current candidate.

Now $K(M_{i})= K(x^i)+O(\log n)$ 
and 
$\log m_{i} =  n-i$, so the MDL code length 
$\Lambda (M_i) = K(x^i) +n-i+O(\log n)$. 
Our MDL algorithm uses a compressor that does not
compress $x^i$ all the way to length 
$K(x^i)$, but codes 
$x^i$ self-delimitingly at $0.9i$ bits, 
that is, it compresses $x^i$
by 10\%.
Thus, the MDL code length is $0.9i+ \log m_{i} 
= 0.9i+  n-i  = n-0.1i $ for every
contemplated model $M_i$ ($i=0, \frac{1}{3}n, \frac{2}{3}n, n$). 
The next equalities hold again up to $O(\log n)$ additive terms.

\begin{itemize}
\item
The MDL code length
of the initial candidate model $M_0$ is $n$. 
The randomness deficiency $\delta (x|M_0) = n - K(x|M_0) =
\frac{1}{3}n$. The last equality holds
since clearly $K(x|M_0)=K(x|n)= \frac{2}{3}n$.
\item
For the contemplated model $M_{n/3}$ we obtain the following.
The MDL code  length for model $M_{n/3}$ 
is $n-n/30$. 
The randomness deficiency
$\delta (x|M_{n/3})= \log m_{n/3} - K(x| M_{n/3}) = \frac{2}{3}n
-  K(v|n)-K(w|n) = \frac{1}{9}n$. 
\item
For the contemplated model $M_{2n/3}$  we obtain the following.
The MDL code length is $n-2n/30$.
The randomness deficiency is
$\delta (x|M_{2n/3})=\log m_{2n/3} - K(x| M_{2n/3}) = \frac{1}{3}n - K(w|n) = \frac{2}{9}n$.
\end{itemize}

Thus, our MDL algorithm initializes with candidate model $M_0$, then
switches to candidate $M_{n/3}$ since this model decreases
the MDL code length by $n/30$. Indeed,
$M_{n/3}$ is a much better model than $M_0$, since it
decreases the randomness deficiency by a whopping $\frac{2}{9}n$.
Subsequently, however, the MDL process switches to candidate
model $M_{2n/3}$ since it decreases the MDL code length greatly again,
by $n/30$. But $M_{2n/3}$ is a much worse model
than the previous candidate $M_{n/3}$, since it increases
the randomness deficiency again greatly by  $\frac{1}{9}n$.
\end{proof}
\begin{remark}
\rm
By Theorem~\ref{theo.approxim} we know that 
if in the process of MDL estimation
by a sequence of significantly decreasing MDL codes
a candidate model is represented by its shortest program,
then the following candidate model which improves the MDL code
is actually a model of at least as good fit as the preceding one.
Thus, if in the example used in the proof above we encode the
models at shortest code length, we obtain MDL code lengths
$n$ for $M_0$, $K(u)+\frac{2}{3}n= \frac{7}{9}n$ for $M_{n/3}$, and
$K(u)+K(v)+ \frac{1}{3}n= \frac{8}{9}n$ for $M_{2n/3}$. Hence the MDL estimator
using shortest model code length changes candidate model $M_0$
for  $M_{n/3}$, improving the MDL code length by $\frac{2}{9}n$ and the
randomness deficiency by $\frac{2}{9}n$. However, and correctly,
it does not change candidate model  $M_{n/3}$ for  $M_{2n/3}$,
since that would increase the MDL code length by $\frac{1}{9}n$. It so
prevents, correctly, to increase the randomness deficiency by $\frac{1}{9}n$. 
Thus, by the cited theorem, the oscillating randomness deficiency
in the MDL estimation process in the proof above can only
arise in cases where the consecutive candidate models are not
coded at minimum cost while the corresponding two-part MDL code
lengths are decreasing.
\end{remark}

\section{Inferring a Grammar (DFA) From Positive Examples}
\label{sect.multi}

Assume that we want to infer
a language, given a set of positive examples (elements of the language)
$D$.
For convenience we restrict the question to inferring
a language 
$M = M' \bigcup \{\#d\}$ with $M' \subseteq \{0,1\}^n$. 
We can view this as inferring the slice $L^n$ (corresponding to $M'$)
of the target language $L$ consisting of all words of length $n$ in the target
language. Since $D$ consists of a subset of positive examples of $M'$
we have $D \sqsubset M$.
To infer a language $M$ from a set of positive examples $D \sqsubset M$
is, of course, a much more natural situation
than to infer a language from a singleton $x$ 
as in the previous section. Note that the complexity $K(x)$ of a singleton
$x$ of length $n$ cannot exceed $n + O(\log n)$, while the
complexity of a language of which $x$ is an element can rise to $2^n + O( \log n)$. 
In the multiple data sample
setting $K(D)$ can rise to $2^n + O( \log n)$, just as $K(M)$ can. 
That is, the description of $n$ takes $O(\log n)$ bits and the description of 
the characteristic sequence of a subset of $\{0,1\}^n$ may take $2^n$ bits,
everything self-delimitingly.
So contrary to the singleton datum case, in principle 
models $M$ of every possible model complexity can be inferred
depending on the data $D$ at hand. An obvious example is $D=M-\{\#d\}$.
Note that the cardinality of $D$ plays a role here, since the complexity
$K(D|n) \leq \log {{2^n} \choose d} + O(\log d)$ with equality for
certain $D$. 
A traditional and well-studied problem in this setting is
to infer a grammar from a language example.

The field of grammar induction studies among other things
a class of algorithms
that aims at constructing a grammar by means of incremental
compression of the data set represented by the digraph 
of a deterministic finite automaton (DFA) accepting the data set. This digraph
can be seen as a model for the data set.
Every word in the data set is represented as a path in the digraph
with the symbols either on the edges or on the nodes. The learning
process takes the form of a guided incremental compression of the
data set by means of merging or clustering of the nodes in the
graph. None of these algorithms explicitly makes an estimate of
the data-to-model code. Instead they use heuristics to guide
the model reduction. After a certain number of computational steps
 a proposal for a grammar
can be constructed from the current state of the compressed graph.
Examples of such algorithms are SP \cite{Wolff:03-19-193,
DBLP:journals/ngc/Wolff95}, EMILE \cite{ICGI:AdrVer2002},
ADIOS
\cite{solan05languageLearningPNAS}, and a number of DFA induction
algorithms, such as ``Evidence Driven State Merging'' (EDSM),
\cite{ICGI:LanPeaPri98,ECMLPKDD/CFG03}. Related compression-based theories
and applications appear in \cite{LCLMV04,CV07}.
Our results (above and below) do not imply that compression
algorithms improving the MDL code of DFAs
 can never work on real life data sets. There is considerable
empirical evidence that there are situations in which they
do work. In those cases specific properties of a restricted class of
languages or data sets must be involved. 

Our results are 
applicable to the common digraph simplification
techniques used in grammar inference.
The results hold equally for
algorithms that use just positive examples, just negative examples, or both,
using any technique (not just digraph simplification).

\begin{definition}
A DFA $A=(S,Q,q_0,t,F)$, where $S$ is a finite set of {\em input symbols},
$Q$ is a finite set of {\em states}, $t: Q \times S \rightarrow Q$
is the {\em transition function}, $q_0 \in Q$ is the
{\em initial state}, and $F \subseteq Q$ is a set
of {\em final states}.
\end{definition}

The DFA $A$ is started in the initial state $q_0$. 
If it is in state $q \in Q$ and receives input symbol $s \in S$
it changes its state to $q' = t(q,s)$. If the machine after zero or more
input symbols, say $s_1, \ldots , s_n$, is driven to a state $q \in F$ 
then it is said to {\em accept} the word $w=s_1 \ldots s_n$, otherwise
it {\em rejects} the word $w$. The {\em language accepted} by $A$
is $L(A)= \{w: w$ is accepted by $A\}$. We denote $L^n(A)= L(A) \bigcap \{0,1\}^n$.

We can effectively enumerate the DFAs as $A_1, A_2 , \ldots$ in
lexicographic length-increasing order. This enumeration we call
the {\em standard enumeration}.

The first thing we need to do is to show that all laws that hold
for finite-set models also hold for DFA models, so all theorems, lemmas,
and remarks above, both positive and negative, apply.
To do so, we show that for every data sample $D \subseteq \{0,1\}^n$
and a contemplated finite set model for it, there
is an almost equivalent DFA.
                                                                                   
\begin{lemma}\label{prop.1}
Let $d=|D|$, $M'=M-\{\#d\}$ and $m=|M'|$. 
For every $D \subseteq M' \subseteq \{0,1\}^n$ there is
a DFA $A$ with $L^n(A)=M'$ such that
$K(A,n) \leq K(M')+ O(1)$ (which implies $K(A,d,n) \leq K(M)+ O(1)$), and
$\delta(D \mid M) \leq \delta(D \mid A,d,n) +O(1)$.
\end{lemma}

\begin{proof}
Since $M'$ is a finite set of binary strings, there is a DFA
that accepts it, by elementary formal language theory.
Define DFA $A$ such that $A$ is the first DFA in the standard
enumeration for which $L^n(A)=M'$. (Note that we can infer $n$ from both 
$M$ and $M'$.)
Hence, $K(A,n) \leq K(M')+O(1)$ and $K(A,d,n) \leq K(M)+O(1)$.
Trivially, $\log {m \choose d} = \log {{|L^n(A)|} \choose d}$
and $K(D \mid  A,n) \leq K(D \mid M')+O(1)$,
 since $A$ may have information about $D$ beyond $M'$. 
This implies $K(D \mid  A,d,n) \leq K(D \mid M)+O(1)$, so that
$\delta(D \mid M) \leq \delta(D \mid A,d,n)+O(1)$. 
\end{proof}

Lemma~\ref{prop.2} is the converse of Lemma~\ref{prop.1}:
for every data sample $D$ and a contemplated
DFA model for it,
there is a finite set model for $D$ that has no worse complexity,
randomness deficiency, and worst-case data-to-model code for $D$,
up to additive logarithmic precision.

    \begin{lemma}\label{prop.2}
    Use the terminology of Lemma~\ref{prop.1}.
    For every $D  \subseteq  L^n(A) \subseteq \{0,1\}^n$, 
    there is a model $M \sqsupset D$
    such that $\log {m \choose d} =  \log {{|L^n(A)| } \choose d}$, 
    $K(M') \leq  K(A,n)+O(1)$ (which implies
    $K(M) \leq  K(A,d,n)+O(1)$), and
    $\delta(D \mid M) \leq \delta(D \mid A,d,n) -O(1)$.
    \end{lemma}
                                                                                   
\begin{proof}
Choose $M' =L^n(A)$. Then, 
$\log {m \choose d} =  \log {{| L^n(A)|} \choose d}$ 
and both $K(M') \leq  K(A,n)+O(1)$ and $K(M) \leq  K(A,d,n)+O(1)$.
Since also  $K(D \mid  A,d,n) \leq K(D \mid M)+O(1)$, 
since $A$ may have information about $D$ beyond $M$, we have
$\delta(D \mid A,d,n) \geq \delta(D \mid M)+O(1)$.
\end{proof}

\subsection{MDL Estimation}
To analyze the MDL estimation for DFAs, given a data sample, we first
fix details of the code. For the model code, the coding of the DFA,
we encode as follows. Let $A=(Q,S,t,q_0,F)$ with $q=|Q|$, $s=|S|$,
and $f=|F|$. 
By renaming of
the states we can always take care that $F \subseteq Q$ are the 
last $f$ states of $Q$. There 
are $q^{sq}$ different possibilities for $t$,  
$q$ possibilities for $q_0$, and $q$ possibilities for $f$.
Altogether, for every choice of $q,s$ there are
$\leq q^{qs+2}$ distinct DFAs, some of which may accept the same languages.

{\bf Small Model Cost but Difficult to Decode:}
We can enumerate the DFAs by setting $i: = 2, 3, \ldots ,$ and for every
$i$ consider all partitions $i=q + s$ to two positive
integer summands, and for every particular choice of $q,s$ considering every
choice of final states, transition function, and initial state.
This way we obtain a standard  enumeration $A_1, A_2, \ldots$ of all DFAs,
and, given the index $j$ of a DFA $A_j$ we can retrieve the particular
DFA concerned, and for every $n$ we can find $L^n(A_j)$. 

{\bf Larger Model Cost but Easy to Decode:}
We encode a DFA $A$ with $q$ states and $s$ symbols self-delimitingly by 
\begin{itemize}
\item
The encoding of the number of symbols $s$ in self-delimiting format
in $\lceil \log s \rceil + 2 \lceil \log \log s \rceil +1$ bits;
\item
The encoding of the number of states $q$ in self-delimiting format
in $\lceil \log q \rceil + 2\lceil \log \log q \rceil +1$ bits;
\item
The encoding of the set of final states $F$ by indicating
that all states numbered $q-f, q-f+1, q$ are final states,
by just giving $q-f$ in $\lceil \log q \rceil$ bits;
\item
The encoding of the  initial state $q_0$ by giving its index
in the states $1, \ldots , q$, in $\lceil \log q \rceil$ bits; and
\item
The encoding of the transition function $t$ in lexicographic
order of $Q \times S$ in $\lceil \log q \rceil$ bits per transition, 
which takes
$qs \lceil \log q \rceil$ bits altogether. 
\end{itemize}
Altogether, this encodes $A$ in a self-delimiting format in
$(qs+3)  \lceil \log q \rceil +  2 \lceil \log \log q \rceil
+\lceil \log s \rceil + 2 \lceil \log \log s \rceil +O(1) \approx 
(qs+4) \log q + 2 \log s$ bits. Thus, we reckon the model cost of
a $(q,s)$-DFA as $m(q,s)=(qs+4) \log q + 2 \log s$ bits.
This cost has the advantage that it is easy to decode and
that $m(q,s)$ is an easy function of $q,s$. We will assume this model cost.

{\bf Data-to-model cost:}
Given a DFA model $A$, the word length $n$
in $\log n + 2 \log \log n$ bits which we simplify to $2 \log n$ bits,
and the size $d$ of the data sample $D \subseteq \{0,1\}^n$,
we can describe $D$ by its index $j$ in the set of $d$ choices out of $l=L^n(A)$
items, that is, up to rounding upwards, $\log {l \choose d}$ bits.
For $0 < d \leq l/2$ this can be estimated by  $l H(d/l) - \log l/2 +O(1) 
\leq \log {l \choose d} \leq l H(d/l)$, where $H(p)= p \log 1/p + 
(1-p) \log 1/(1-p)$
($0 < p < 1$) is Shannon's entropy function. 
For $d=1$ or $d=l$ we set the data-to-model cost to $1 + 2 \log n$, 
for $1 < d \leq l/2$ 
we set it to $2 \log n + l H(d/l)$ (ignoring the possible 
saving of a $\log l/2$ term), and for $l/2 <d <l$
we set it to the cost of $d'=l-d$. This reasoning brings us to the following
MDL cost of a data sample $D$ for DFA model $A$:

\begin{definition}
\rm
The {\em MDL code length} of a data sample $D$ of $d$
strings of length $n$, given $d$, for a DFA model $A$ such that 
$D \subseteq L^n(A)$ denoting $l=|L^n(A)|$,
is given by
\[
MDL (D,A|d)= (qs+4) \log q + 2 \log s + 2 \log n + l H(d/l).
\]   
If $d$ is not given we write $MDL (D,A)$.
\end{definition}

\subsection{Randomness Deficiency Estimation}
Given data sample $D$ and DFA $A$ with 
$D \subseteq L^n(A) \subseteq \{0,1\}^n$,
we can estimate the randomness deficiency. 
Again, use $l= L^n(A)$ and $d=|D|$.
By \eqref{eq:randomness-deficiency}, the randomness deficiency is
\[
\delta (D \mid A,d,n) = \log  {l \choose d} - K(D \mid A,d,n). 
\]
Then, substituting the estimate for $\log {l \choose d}$ from the previous
section, up to logarithmic additive terms,
\[
\delta (D \mid A,d,n) =  l H(d/l) -  K(D \mid A,d,n). 
\]
Thus, by finding a computable upper bound for $K(D \mid A,d,n)$,
we can obtain a computable lower bound on the randomness
deficiency $\delta (D \mid A,d,n)$ that expresses the fitness
of a DFA model $A$ with respect to data sample $D$.

\subsection{Less MDL Code Length Doesn't Mean Better Model}
\label{sect.lmdl}
The task of finding the smallest {DFA} consistent with a set of
positive examples is trivial. This is the universal DFA accepting 
every example (all of $\{0,1\}^n$). Clearly, such a
universal DFA will in many cases have a poor generalization error
and randomness deficiency. As we have seen, optimal randomness deficiency 
implies an optimal fitting model to the data sample. It is to be expected
that the best fitting model gives the best generalization error in the case
that the future data are as typical to this model as the data sample is.
We show that the
randomness deficiency behaves independently of the MDL code, in the sense
that the randomness deficiency can either grow or shrink with a
reduction of the length of the MDL code. 

We show this by example.
Let the set $D$ be a sample set consisting of 50\% of all binary
strings of length $n$ with an even number of 1's. Note, that the 
number of strings with an even number of 1's equals the number of strings
with an odd number of 1's, so $d=|D|= 2^{n}/4$.
Initialize with a DFA $A$ such that $L^n(A)=D$. We can obtain
$D$ directly from $ A,n$, so  we have $K(D \mid A,n)=O(1)$, and since $d=l$
($l=|L^n(A)|$) we have $\log {l \choose d} =0$, so that
altogether $\delta (D \mid A,d,n)= -O(1)$,
while $MDL(D,A) = MDL(D,A|d)+O(1)
= (qs+4) \log q + 2 \log s + 2 \log n +O(1)=
(2q+4) \log q + 2 \log n +O(1)$, since $s=2$. 
(The first equality follows since we can obtain $d$ from $n$.
We obtain a negative constant randomness deficiency which we take
to be as good as 0 randomness deficiency. All arguments hold up to
an $O(1)$ additive term anyway.) 
Without loss of generality we can assume that the MDL algorithm
involved works by splitting or merging nodes of the digraphs
of the produced sequence of candidate DFAs. But the argument
works for every MDL algorithm, whatever technique it uses.

{\em Initialize:} Assume that we start our MDL estimation
with the trivial DFA $A_0$ that literally encodes
all $d$ elements of $D$ as a binary directed tree with $q$ nodes. 
Then, $2^{n-1} - 1 \leq q \leq 2^{n+1}-1$, which yields
\begin{align*}
&MDL (D,A_0) \geq 2^n n
\\& \delta (D \mid A_0,d,n) \approx 0. 
\end{align*}
The last approximate equality holds since $d=l$, and hence $\log {l \choose d} =0$
and $K(D \mid A_0,d,n)=O(1)$.
Since the randomness deficiency
$ \delta (D \mid A_0,d,n) \approx 0$,  it follows that $A_0$ is 
a best fitting model for $D$. Indeed, it represents all conceivable
properties of $D$ since it literally encodes $D$. However, $A_0$
does not achieve the optimal MDL code.

{\em Better MDL estimation:}
In a later MDL estimation we improve the MDL code by inferring
the parity DFA $A_1$ with two states ($q=2$) that checks the 
parity of 1's in a sequence. Then,
\begin{align*}
&MDL (D,A_1) \leq 8 + 2\log n+ \log {{2^{n-1}} \choose {2^{n-2}}} \approx
 2^{n-1} - \frac{1}{4}n
\\& \delta (D \mid A_1,d,n) = \log {{2^{n-1}} \choose {2^{n-2}}} 
- K(D \mid A_1,d,n)
\\&\approx 2^{n-1} - \frac{1}{4}n- K(D \mid A_1,d,n)
\end{align*}
We now consider two different instantiations of $D$, denoted
as $D_0$ and $D_1$. The first one is regular data, and the
second one is random data.

{\bf Case 1, regular data:}
Suppose $D=D_0$ consisting of the lexicographic first 50\% of all $n$-bit 
strings with an even number of occurrences of 1's.
Then $K(D_0 \mid A_1,d,n)=O(1)$ and 
\[
\delta (D_0 \mid A_1,d,n) = 2^{n-1} - O(n).
\]
In this case, even though DFA $A_1$ has a much better MDL code than DFA $A_0$
it has nonetheless a much worse fit since its randomness deficiency
is far greater.  

{\bf Case 2, random data:}
Suppose $D$ is equal to $D_1$, where $D_1$ is a random subset consisting of 50\%
of the $n$-bit strings with even number of occurrences of 1's.
Then, 
$K(D_1 \mid A_1,d,n) = \log {{2^{n-1}} \choose {2^{n-2}}}+O(1)
\approx 2^{n-1} -\frac{1}{4} n$, and 
\[
\delta (D_1 \mid A_1,d,n) \approx 0.
\]
In this case, DFA $A_1$ has a much better MDL code than DFA $A_0$,
and it has equally good fit since both randomness deficiencies
are about 0.

\begin{remark}
We conclude that improved MDL estimation of DFAs for multiple data
samples doesn't necessarily result in better models, but can do so
nonetheless. 
\end{remark}

\begin{remark}[Shortest Model Cost]\label{rem.smc}
\rm
By Theorem~\ref{theo.approxim} we know that if, in the process of MDL estimation 
by a sequence of significantly decreasing MDL codes, 
a candidate DFA is represented by its shortest program,
then the following candidate DFA which improves the MDL estimation
is actually a model of at least as good fit as the preceding one.
Let us look at an Example:
Suppose we start with DFA $A_2$ that accepts all strings
in $\{0,1\}^*$. In this case we have $q=1$ and 
\begin{align*}
&MDL (D_0,A_2) = \log {{2^n} \choose {2^{n-2}}} +O(\log n)
\\& \delta (D_0 \mid A_2,d,n) =   \log {{2^n} \choose {2^{n-2}}}-O(1).
\end{align*}
Here $ \log {{2^n} \choose {2^{n-2}}} = 2^n H(\frac{1}{4})-O(n) 
\approx \frac{4}{5} \cdot 2^{n}
-O(n)$, since $H(\frac{1}{4}) \approx \frac{4}{5}$.
Suppose the subsequent candidate DFA is the parity machine $A_1$.
Then,
\begin{align*}
&MDL (D_0,A_1)= \log {{2^{n-1}} \choose {2^{n-2}}} +O(\log n)
\\&
\delta (D_0 \mid A_1,d,n) \approx 
 \log {{2^{n-1}} \choose {2^{n-2}}} 
 - O(1),
\end{align*}
since $K(D_0 \mid A_1,d,n)=O(1)$. Since
 $\log {{2^{n-1}} \choose {2^{n-2}}} 
=2^{n-1}-O(n)$, we have
$MDL (D_0,A_1 ) \approx \frac{5}{8} MDL (D_0,A_2 )$, 
and 
$\delta (D_0 \mid A_1,d,n) \approx 
 \frac{5}{8} \delta (D_0 \mid A_2,d,n)$.
Therefore, the improved MDL cost from model $A_2$ to
model $A_1$ is accompanied by an improved model fitness since
the randomness deficiency decreases as well. This 
is forced by  Theorem~\ref{theo.approxim}, since 
both DFA $A_1$ and DFA $A_2$ have $K(A_1),K(A_2)= O(1)$.
That is, the DFAs are represented and penalized according
to their shortest programs (a fortiori of length $O(1)$) and therefore
improved MDL estimation increases the fitness of the
successive DFA models significantly. 
\end{remark}

\appendix
\subsection{Appendix: Preliminaries}
\label{sect.prel}
\subsubsection{Self-delimiting Code}
A binary string $y$
is a {\em proper prefix} of a binary string $x$
if we can write $x=yz$ for $z \neq \epsilon$.
 A set $\{x,y, \ldots \} \subseteq \{0,1\}^*$
is {\em prefix-free} if for every pair of distinct
elements in the set neither is a proper prefix of the other.
A prefix-free set is also called a {\em prefix code} and its
elements are called {\em code words}.
As an example of a
prefix code, 
encode the source word $x=x_1 x_2 \ldots x_n$ by the code word
\[ \overline{x} = 1^n 0 x .\]
This prefix-free code
is called {\em self-delimiting}, because there is fixed computer program
associated with this code that can determine where the
code word $\bar x$ ends by reading it from left to right without
backing up. This way a composite code message can be parsed
in its constituent code words in one pass, by the computer program.
Since we use the natural numbers and the binary strings interchangeably,
the notation $|\bar x|$ where $x$ is ostensibly an integer means the length
in bits of the self-delimiting code of the $x$th binary string.
On the other hand, the notation $\overline{|x|}$ where $x$ is ostensibly a binary
string means the self-delimiting code of the length $|x|$ of the binary string
$x$.
Using this code we define
the standard self-delimiting code for $x$ to be
$x'=\overline{|x|}x$. It is easy to check that
$|\overline{x} | = 2 n+1$ and $|x'|=n+2 \log n +1$.
Let $\langle \cdot \rangle$ denote a standard invertible
effective one-to-one code from ${\cal N} \times {\cal N}$
to a subset of ${\cal N}$.
For example, we can set $\langle x,y \rangle = x'y$
or $\langle x,y \rangle = \bar xy$.
We can iterate this process to define
$\langle x , \langle y,z \rangle \rangle$,
and so on.
                                                                                
\subsubsection{Kolmogorov Complexity}
For precise definitions, notation, and results see the textbook \cite{LiVi97}.
Informally, the Kolmogorov complexity, or algorithmic entropy, $K(x)$ of a
string $x$ is the length (number of bits) of a shortest binary
program (string) to compute
$x$ on a fixed reference universal computer
(such as a particular universal Turing machine).
Intuitively, $K(x)$ represents the minimal amount of information
required to generate $x$ by any effective process.
The conditional Kolmogorov complexity $K(x | y)$ of $x$ relative to
$y$ is defined similarly as the length of a shortest program
to compute $x$, if $y$ is furnished as an auxiliary input to the
computation.
For technical reasons we use a variant of complexity,
so-called prefix complexity, which is associated with Turing machines
for which the set of programs resulting in a halting computation
is prefix free.
We realize prefix complexity by considering a special type of Turing
machine with a one-way input tape, a separate work tape,
and a one-way output tape. Such Turing
machines are called {\em prefix} Turing machines. If a machine $T$ halts
with output $x$
after having scanned all of $p$ on the input tape,
but not further, then $T(p)=x$ and
we call $p$ a {\em program} for $T$.
It is easy to see that
$\{p : T(p)=x, x \in \{0,1\}^*\}$ is a {\em prefix code}.

Let $T_1 ,T_2 , \ldots$ be a standard enumeration
of all prefix Turing machines with a binary input tape,
for example the lexicographic length-increasing ordered syntactic
prefix Turing machine descriptions, 
and let $\phi_1 , \phi_2 , \ldots$
be the enumeration of corresponding functions
that are computed by the respective Turing machines
($T_i$ computes $\phi_i$).
These functions are the {\em partial recursive} functions
or {\em computable} functions (of effectively prefix-free encoded
arguments). The prefix (Kolmogorov) complexity
of $x$ is the length of the shortest binary program
from which $x$ is computed.
For the development of the theory we
require
the Turing machines to use {\em auxiliary} (also
called {\em conditional})
information, by equipping the machine with a special
read-only auxiliary tape containing this information at the outset.
                                                                                
One of the main achievements of the theory of computation
is that the enumeration $T_1,T_2, \ldots$ contains
a machine, say $U=T_u$, that is computationally universal in that it can
simulate the computation of every machine in the enumeration when
provided with its index:
    $U(\langle y, \bar{i}p)  = T_i (\langle y,p\rangle)$
    for all $i,p,y$.
    We fix one such machine and designate it as the {\em reference universal
    prefix Turing machine}.

\begin{definition}\label{def.KolmK}
    Using this universal machine we define the {\em prefix (Kolmogorov) complexity}
                  \begin{equation}\label{eq.KC}
    K (x \mid y) = \min_q \{|q|: U(\langle y,q\rangle)=x \},
                  \end{equation}
the {\em conditional version} of the prefix
Kolmogorov complexity of $x$
given $y$ (as
auxiliary information).
The unconditional version is set to  $K(x)=K(x  \mid \epsilon)$.
\end{definition}

In this paper we use the prefix complexity
variant of Kolmogorov complexity only for convenience; the plain
Kolmogorov complexity without the prefix property would do just as well.
  The functions $K( \cdot)$ and $K( \cdot \mid  \cdot)$,
though defined in terms of a
particular machine model, are machine-independent up to an additive
constant
 and acquire an asymptotically universal and absolute character
through Church's thesis, that is, from the ability of universal machines to
simulate one another and execute any effective process.
  The Kolmogorov complexity of an individual object was introduced by
Kolmogorov \cite{Ko65} as an absolute
and objective quantification of the amount of information in it.
The information theory of Shannon \cite{Sh48}, on the other hand,
deals with {\em average} information {\em to communicate}
objects produced by a {\em random source}.
  Since the former theory is much more precise, it is surprising that
analogues of theorems in information theory hold for
Kolmogorov complexity, be it in somewhat weaker form.
An example is the remarkable {\em symmetry of information} property.
Let $x^*$  denote the shortest prefix-free program 
for a finite string $x$,
or, if there are more than one of these, then $x^*$ is the first
one halting in a fixed standard enumeration of all halting programs.
It follows that $K(x)=|x^*|$.
Denote $K(x,y)=K(\langle x,y \rangle)$. Then,
\begin{align}\label{eq.soi}
K(x,y) & = K(x)+K(y \mid x^*) + O(1) \\
& = K(y)+K(x \mid y^*)+O(1) .
\nonumber
\end{align}
                                                                                
\subsubsection{Precision}
It is customary in this area to use ``additive constant $c$'' or
equivalently ``additive $O(1)$ term'' to mean a constant,
accounting for the length of a fixed binary program,
independent from every variable or parameter in the expression
in which it occurs.

\subsection{Appendix: Structure Functions and Model Selection}
\label{sect.formal}
We summarize a selection of the results in \cite{VV02}.
There, the data sample $D$ is a singleton set $\{x\}$.
The results extend  to the multiple data sample case
in the straightforward way.

(i) The MDL code length $\lambda_D (\alpha)$ with $D \subseteq \{0,1\}^n$
and $d=|D|$
can assume essentially every possible relevant shape $\lambda (\alpha)$
as a function of the 
maximal model complexity $\alpha$ 
that is allowed up to an additive $O(\log dn)$ term in
argument and value. (Actually, we can take this term
as $O(\log n + \log \log {{2^n} \choose d})$, but since this is
cumbersome we use the larger $O(\log dn)$ term. The difference becomes
large for $2^{n-1} < d  \leq 2^n$.)
 These $\lambda$'s are all integer-valued
nonincreasing functions such that
$\lambda$ is defined on $[0,k]$ where $k=K(D)$,
such that $\lambda(0) \leq \log {{2^n} \choose d}$ and
$\lambda (k)=k$. This is Theorem IV.4 in \cite{VV02} 
for singleton data $x$. There,  $\lambda_x$ is contained in a strip
of width $O(\log n)$ around $\lambda$. For multiple data $D$ ($|D|=d$)
a similar theorem holds up to an 
$O(\log dn)$ 
additive term in both argument and value, that is, the strip
around $\lambda$ in which $\lambda_D$ is situated now has width $O(\log nd)$.
(The strip idea is made precise in
\eqref{eq.multipleeq} below for \eqref{eq.eq}, another result.)
As a consequence, so-called ``nonstochastic'' data $D$
for which $\lambda_D(\alpha)$
stabilizes on $K(D)$ only for large $\alpha$ are common.
                                                                                 
(ii) A model achieving the MDL code length $\lambda_D (\alpha)$,
essentially achieves the best possible fit
$\beta_D (\alpha)$. This is Theorem IV.8 in \cite{VV02} for singleton data
and \eqref{eq.eq} in this paper for multiple data. 
The precise form is:
\begin{eqnarray}\label{eq.multipleeq}
\beta_D (\alpha ) +K(D) &\geq& \min\{ \lambda_D (\alpha '): 
|\alpha' - \alpha| 
= O(\log dn)\}
\nonumber
 \\&&-O(\log dn),
\\ \beta_D (\alpha ) +K(D) &\leq &\max\{ \lambda_D (\alpha '): 
|\alpha' - \alpha| 
\nonumber
 = O(\log dn)\} \\&&+O(\log dn),
\nonumber
\\ \lambda_D (\alpha ) -K(D) &\geq& \min\{ \beta_D (\alpha '): 
|\alpha' - \alpha| 
\nonumber
= O(\log dn)\} \\&&-O(\log dn),
\nonumber
\\ \lambda_D (\alpha ) -K(D) &\leq & \max\{ \beta_D (\alpha '): 
|\alpha' - \alpha| 
\nonumber
= O(\log dn)\} \\&& +O(\log dn),
\nonumber
\end{eqnarray}
with  $0 \leq \alpha \leq K(D)$ and $O(\log dn) \leq \alpha' \leq K(D)$.

(iii) As a consequence of (i) and (ii), the best-fit function $\beta_D$
can assume essentially every possible relevant shape as a function of the
contemplated maximally allowed model complexity $\alpha$.

From the proof of Item (ii),
we see  that, given the data sample $D$,
for every finite set $M\sqsupset D$, of complexity at most
$\alpha+O(\log dn)$ and  minimizing $\Lambda(M)$,
we have
$\delta(D \mid M)\le \beta_D(\alpha)+O(\log dn)$.
Ignoring $O(\log dn)$ terms, at every complexity
level $\alpha$, every best model
at this level witnessing $\lambda_D(\alpha)$ is
also a best one with respect to
typicality \eqref{eq.eq}. This explains why it is worthwhile to
find shortest two-part descriptions $\lambda_D(\alpha)$
for the given data sample $D$: this is the single known
way to find an $M\sqsupset D$ with respect to
which $D$ is as typical as possible at model complexity level $\alpha$.
Note that the set $\{\pair{D,M,\beta}\mid D \subseteq  M,\ \delta(D \mid
M)<\beta\}$ is not enumerable so we are not able to generate such
$M$'s directly, \cite{VV02}.
                                                                                 
The converse is not true: not every model
(a finite set) witnessing
$\beta_D(\alpha)$ also witnesses $\lambda_D(\alpha)$.
For example, let $D=\{x\}$ with $x$  a
string of length $n$ with $K(x) \geq n$. Let $M_1=
\{0,1\}^n\cup\{y0 \ldots 0\}$ 
(we ignore the $\{\#1\}$
set giving the data sample cardinality since $D$ is a singleton set),
where $y$ is a string of length $\frac{n}{2}$ such
that $K(x,y)\ge \frac{3n}{2}$
and let $M_2= \{0,1\}^n$.
Then
both $M_1,M_2$ witness
$\beta_D(\frac{n}{2}+O(\log n))=O(1)$
but
$\Lambda(M_1)=\frac{3n}{2}+O(\log n)\gg
\lambda_D(\frac{n}{2}+O(\log n))=n+O(\log
n)$
while $\Lambda (M_2) = n+O(\log n)$.

\section*{Biography}
{\sc Pieter Adriaans} received his Ph.D. from the University of Amsterdam (1992).
He and his business partner, Dolf Zantinge, founded  the software 
developer Syllogic B.V. in 1989, and sold the company 
to Perot Systems Corporation in 1997. 
Adriaans is Professor of Computer Science at the University of Amsterdam since 1997.
He serves as editor of the Handbook of Philosophy of Information, a project of 
Elseviers Science Publishers, and is a member of the ICGI (International Conference 
on Grammar Induction) steering committee. He is adviser of Robosail Systems, 
a company that manufactures and sells self-learning autopilots, 
as well as senior research adviser for Perot Systems Corporation. He has worked on 
learning, 
grammar induction,
philosophy of information, and
information and art. 
 He holds several patents on adaptive systems 
management and on a method for automatic composition of music using grammar induction 
techniques. Adriaans acted as project leader for various large international 
research and development projects: amongst others, the development of distributed database management 
software in cooperation with IBM and Prognostic and Health management for the 
Joint Strike Fighter. He wrote papers and books on 
topics related to both computer science and philosophy, 
including a book on systems analysis and books on client/server 
and distributed databases as well as data mining. He composes and plays rock music and
is an avid painter. 
In 2006 he had an overview exhibition showing the harvest of forty years of 
painting.

{\sc Paul M.B. Vit\'anyi} received his Ph.D. from the Free University of 
Amsterdam (1978). He is a  Fellow
at the national research institute for mathematics and computer science (CWI)
in the Netherlands, and Professor of Computer Science
at the University of Amsterdam.  He serves on the editorial boards
of Distributed Computing (until 2003), Information Processing Letters,
Theory of Computing Systems, Parallel Processing Letters,
International journal of Foundations of Computer Science, 
Journal of Computer and Systems Sciences (guest editor),
and elsewhere. He has worked on cellular automata,
computational complexity, distributed and parallel computing,
machine learning and prediction, physics of computation,
Kolmogorov complexity, information theory, and quantum computing,
publishing about 200 research papers and some books. 
He received a knighthood in 2007. Together with Ming Li
they pioneered applications of Kolmogorov complexity
and co-authored ``An Introduction to Kolmogorov Complexity
and its Applications,'' Springer-Verlag, New York, 1993 (second edition 1997,
third edition 2008),
parts of which have been translated into Chinese, Russian, and Japanese.
\end{document}